\documentclass[conference,flushend]{iaria} % (based on IEEEtran.cls)
\usepackage{hyperref}
\title{A Workflow for Map Creation in Autonomous Vehicle Simulations\\}

\author{
  \IEEEauthorblockN{%
    Zubair Islam, Ahmaad Ansari, George Daoud, and Mohamed El-Darieby}
  \IEEEauthorblockA{%
    Faculty of Engineering and Applied Science\\
    Ontario Tech University\\
    Oshawa, Canada\\
    e-mail: {\tt$\lbrace$zubair.islam\,|\,ahmaad.ansari\,|\,george.daoud\,|\,Mohamed.El-Darieby$\rbrace$@ontariotechu.net}
} }

\usepackage{fancyhdr}
\begin{document}
\maketitle
\thispagestyle{fancy}
\begin{abstract}
The fast development of technology and artificial intelligence has significantly advanced Autonomous Vehicle (AV) research, emphasizing the need for extensive simulation testing. Accurate and adaptable maps are critical in AV development, serving as the foundation for localization, path planning, and scenario testing. However, creating simulation-ready maps is often difficult and resource-intensive, especially with simulators like CARLA (CAR Learning to Act). Many existing workflows require significant computational resources or rely on specific simulators, limiting flexibility for developers. This paper presents a custom workflow to streamline map creation for AV development, demonstrated through the generation of a 3D map of a parking lot at Ontario Tech University. Future work will focus on incorporating SLAM technologies, optimizing the workflow for broader simulator compatibility, and exploring more flexible handling of latitude and longitude values to enhance map generation accuracy.
\end{abstract}


\section{Introduction}
With rapid technological advancement, the design and development of Autonomous Vehicles (AVs) has become increasingly common. AVs provide many benefits, such as increased safety, reduced traffic congestion, improved fuel efficiency, and enhanced mobility for individuals who are unable to drive. However, there are many challenges, such as high development costs, lack of detailed regulations, and ethical concerns for decision-making procedures. To address these challenges, research must be conducted on all aspects of an AV, such as path planning, object detection, object avoidance, localization, simulation testing, sensor fusion, and machine learning. Autoware \cite{b1}, an open-source software stack for AVs, provides these functionalities out of the box. For this study, it was selected as the primary software platform due to its widespread adoption in AV research and its robust support for localization and simulation testing. Autoware simplifies development and facilitates integration with simulation platforms. This area of research is especially important for simulation testing due to the accessibility and safety of digitally simulated AVs, allowing for the validation and verification of autonomous systems in a controlled environment without the risks and costs of real-world testing. 

To enable the use of AVs in real-life scenarios, they must first be tested extensively in a simulated environment. A key component of these simulations is high-definition (HD) maps, which provide detailed, centimeter-level accuracy for road layouts, lane markings, and traffic infrastructure. HD maps are essential for localization, perception, and planning, as they allow AVs to understand their position and navigate accurately. The simulated environment can be run in an existing 3D simulation engine, built to support the development and integration of these simulations. Current 3D game engines such as Unity, Unreal Engine, and Godot allow for the creation of highly detailed and interactive virtual environments, which are important for testing the behaviors of AVs. These game engines are utilized to simulate the real world, such as the physics, traffic, pedestrians, and the AV along with its hardware and functionalities. In addition, the maps created for simulation are not only used for virtual testing but also serve as a foundation for real-world deployment. By ensuring accuracy in simulation maps, developers can generate HD maps that are later used by AVs to navigate real-life roads, allowing a seamless transition from testing to deployment.

Currently, there are a few simulators that utilize these game engines. These include Godot, which uses the Godot Engine, AWSIM \cite{b2}, which uses the Unity Engine, and CARLA, \cite{b3} which uses the Unreal Engine. These simulators are built to support user interaction for testing scenarios between AVs and the real world. They are packaged and distributed as easy-to-setup and ready-to-use software designed to aid developers. While these simulators offer customization options to test user-defined environments, the process of developing and integrating new features can be challenging, potentially requiring significant effort to understand and adapt to the tools and workflows specific to each platform. Among these options, AWSIM was selected for this project due to its user-friendly interface and native compatibility with Autoware \cite{b4}, enabling efficient testing and development of AV algorithms. AWSIM supports the project's objectives by enabling detailed testing of localization and vehicle interactions in a controlled and customizable environment.

In this paper, a custom workflow is presented that simplifies the creation of maps that are compatible with AWSIM. Section 2 provides the motivation behind the study. Section 3 reviews related work in HD map generation and simulation-based AV testing. Section 4 details the methodology, describing the tools used, their functionalities, and their integration into the proposed workflow. Section 5 presents the results of implementing the workflow. Section 6 discusses the performance, limitations, and practical advantages of the approach. Finally, Section 7 concludes the paper and outlines directions for future work. While this workflow is tailored for AWSIM, it has the flexibility to be adapted for other simulators, although this potential is not explored in this paper.

\section{Motivation}

At the time of development, AWSIM and Autoware offered only a single simulation map, which represented a large city environment. However, this map lacked an important feature, which was a parking lot. Parking lots are essential for testing real-world AV deployment in low-speed, complex environments where interactions with nearby vehicles are common.

This limitation highlighted the need for a custom parking lot environment. Although documentation existed for creating custom environments in AWSIM, it was difficult to interpret and follow. Through this process, one key requirement was clear, which was that creating a custom environment requires a Lanelet2 OSM file, a PCD (Point Cloud Data) file, and a 3D mesh file. Due to the complexity of the existing documentation, an alternative solution was needed to streamline the map creation process.

Through extensive research and troubleshooting, a custom workflow was developed that utilized multiple tools with different functionalities to generate the required files. By using an OpenStreetMap (OSM) \cite{b5} file and following a series of steps in the workflow, the required files can be exported from the workflow. This allows for the use of a custom environment inside Autoware and AWSIM, enabling simulation testing for any outdoor area available on OSM. OSM was selected as a starting point because it is open-source and has very wide geographic coverage of maps across the globe.

\section{Related Work}
During the development of the workflow, a literature review revealed no prior research specifically addressing map creation using AWSIM. As a result, the workflow was constructed incrementally through extensive Google searches and iterative problem-solving. Each step built upon the previous one, beginning with the extraction of an OSM file, followed by generating a 3D mesh, converting the mesh into a point cloud, and continuing through the necessary processing steps. To emphasize the significance of this workflow and its practical applications, several related studies are discussed below.

In researching methods for creating simulation environments for Autoware and AWSIM, literature was found describing workflows based on different simulation platforms, primarily CARLA and LGSVL. These works often used earlier versions of Autoware or relied on a deprecated simulator, such as LGSVL, making them less applicable to modern systems such as AWSIM.

Feng, Ye, and Angeloudis \cite{b6} proposed a pipeline for Autoware that transforms OSM data into maps compatible with both CARLA and LGSVL. Their workflow begins by converting an OSM file from OpenStreetMap into a 3D model using Blender, a 3D graphics software. This model is then exported in FBX format. Simultaneously, the OSM file is converted into OpenDRIVE format, resulting in both an OpenDRIVE file and a FBX file required by both simulators. Next, a PCD file is generated using CARLA’s PCD recording function, which simulates an AV equipped with a LiDAR sensor that navigates through the environment and captures the PCD data. Finally, the OpenDRIVE file is used to generate a Lanelet2 vector map, enabling integration with Autoware.

Santonato \cite{b7} presents a similar pipeline, though focused solely on CARLA. The process begins by generating an OSM file and converting it into OpenDRIVE format. A plugin called StreetMap for Unreal Engine is then used to render the streets and buildings and to generate the 3D environment. The OpenDRIVE and 3D files are then imported into CARLA to create the simulation map. Lastly, they generate the PCD and Lanelet2 vector map files to be used for Autoware. As in the work by Feng, Ye, and Angeloudis, CARLA is used to record and generate the PCD file, while the OpenDRIVE file is used to create a Lanelet2 vector map.

Both Feng, Ye, and Angeloudis, and Santonato developed workflows capable of creating 3D maps from OSM files. However, their approaches rely on CARLA to generate PCD files. Feng, Ye, and Angeloudis used LGSVL as their primary simulator, which is now deprecated, but depended on CARLA for generating compatible files. Santonato on the other hand, used CARLA as their main simulator, which streamlined the process by keeping all file generation within a single platform. In contrast, the workflow presented in this paper is independent of any specific simulator for file generation. Instead, it utilizes lightweight, open-source tools that are easy to install and do not require running a simulator to produce the necessary files. This flexibility reduces computational overhead and simplifies deployment, particularly for researchers working outside the CARLA ecosystem.

Beyond simulator-based workflows, recent research has explored high-definition (HD) map generation using real-world sensor data. Li et al. \cite{b8} introduced HDMapNet, a deep learning-based framework for generating semantic HD maps online using inputs from cameras and/or LiDAR. Jeong et al. \cite{b9} presented a detailed tutorial for HD map generation using physical vehicles equipped with LiDAR, GNSS, and cameras, involving manual annotation, sensor calibration, and integration with a now-deprecated version of Autoware based on the ROS 1 framework. While both approaches produce high-accuracy maps suitable for deployment, they require significant hardware, real-world data collection, and complex processing pipelines. In contrast, the workflow proposed in this paper is designed specifically for simulation use. It operates entirely offline using publicly available OpenStreetMap data and a set of lightweight, open-source tools to generate Autoware-compatible maps. This makes it especially valuable for early-stage development, academic research, and rapid prototyping within simulation environments like AWSIM, where real-world precision is not critical and accessibility is a key concern.

\section{Methodology}

The workflow is made up of four steps and is shown in Figure 1. The process goes from manually selecting the desired location and inputting that file into an Automated Mapping Docker Container \cite{b10}, shown in Figure 2, which builds the location file into a 3D mesh and extracts the simulated pointcloud. 

\begin{figure}[h]
\centering
\includegraphics[width=\linewidth]{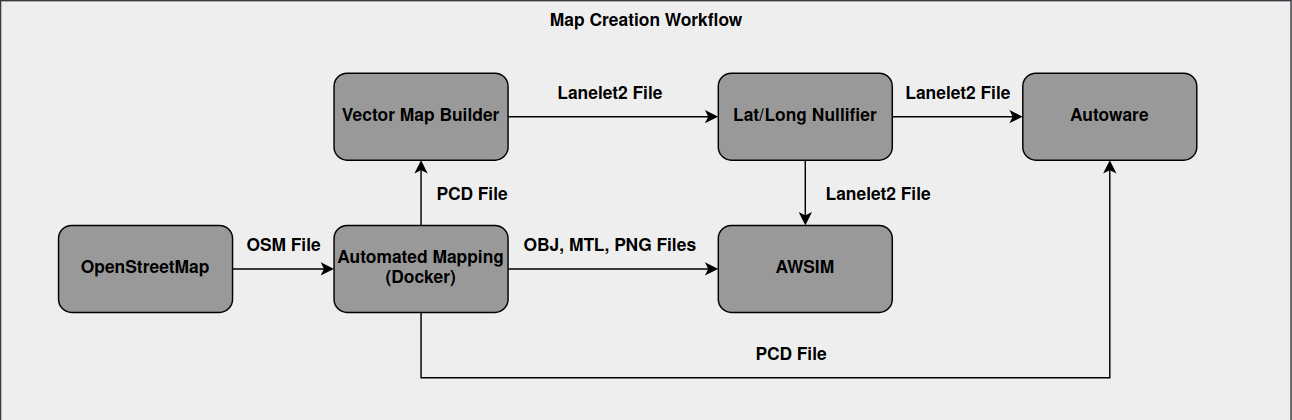}
\caption{Workflow of map creation.}
\end{figure}

\begin{figure}[h]
\centering
\includegraphics[width=\linewidth]{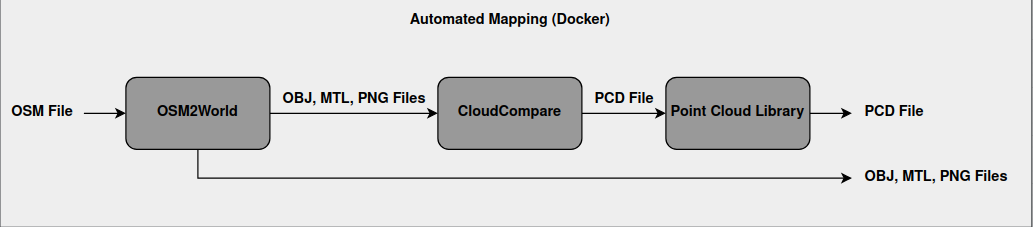}
\caption{Workflow of automated mapping Docker container.}
\end{figure}

After the Automated Mapping Pipeline, the lanes of traffic or parking lots must be manually defined. After this, all the necessary files are generated and can be used for integration with Autoware and AWSIM. Each step requires different tools, each providing different functionalities. The tools will be discussed below and how they were used to develop a map environment for Ontario Tech University’s SIRC parking lot.

\subsection{Functionalities and Usage}

\begin{enumerate}
    \item \textbf{OpenStreetMap (OSM) Selection}
    
    OpenStreetMap \cite{b5} is a resource for getting geospatial data of the world. It allows users to select a certain location to create an environment for. Using this tool, the desired location can be extracted as an OSM file (.osm). This OSM file contains many elements that define the geography and features of the selected location, such as nodes, ways, relations, and tags.
    
    This tool is provided as a website. Using this website, the campus location can be found using its address, providing an aerial view. Using the select tool, the SIRC parking lot can be selected as shown in Figure 3, and exported as an OSM file, containing the geospatial information of this location.

    \begin{figure}[h]
    \centering
    \includegraphics[width=\linewidth]{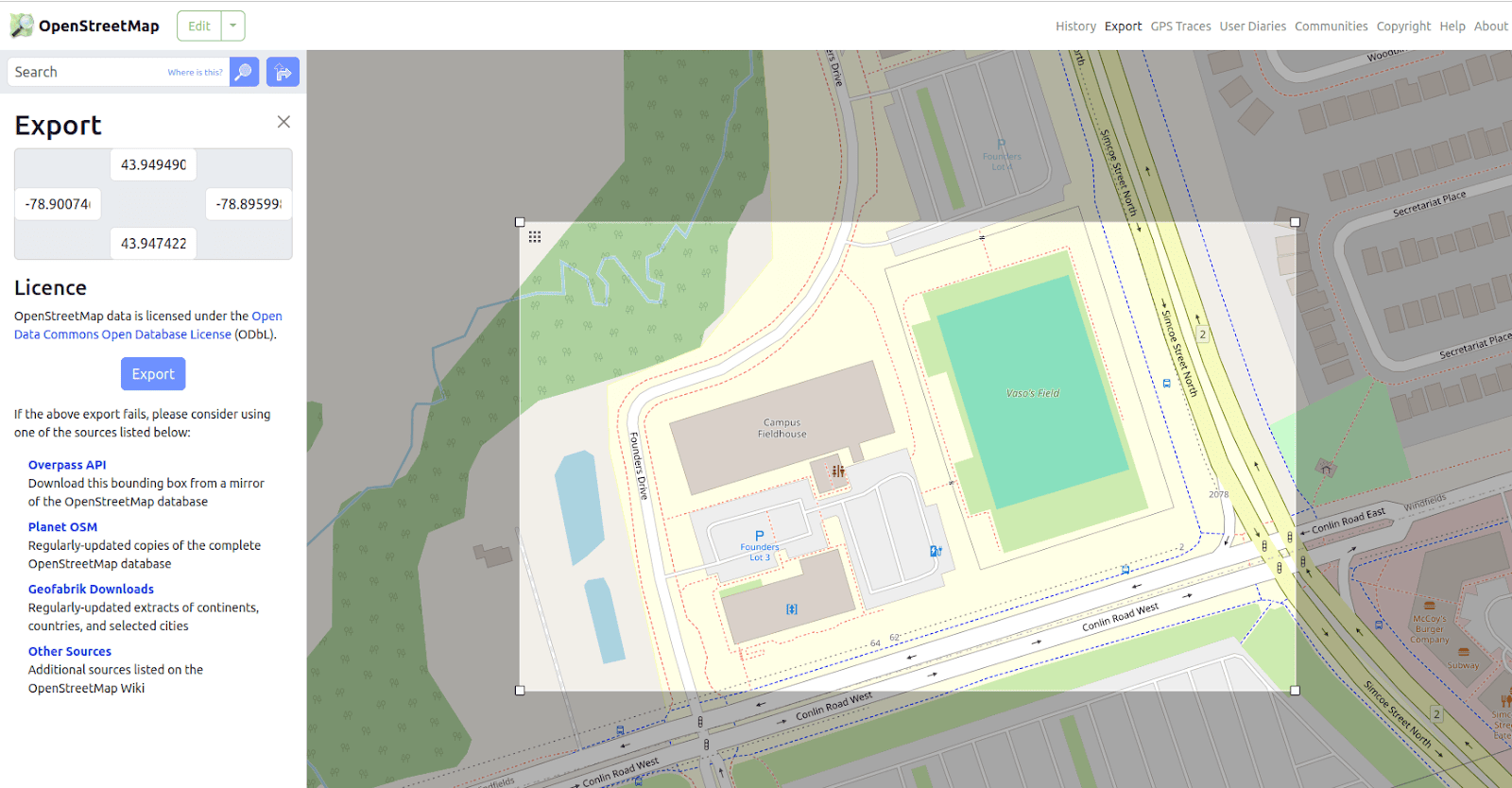}
    \caption{Exporting an OSM file from OpenStreetMap.}
    \end{figure}

    \item \textbf{Automated Mapping Pipeline Docker Container}
        
    This Docker container uses OSM2World \cite{b11} to first generate the 3D model of the map. Next, it uses CloudCompare to generate the Point Cloud Data file, and lastly uses the Point Cloud Library to further process the Point Cloud Data file.

    \begin{enumerate}
        \item \textbf{OSM2World Conversion} 
        
        OSM2World is a conversion tool that generates a 3D mesh based on the provided OSM file. It creates a three-dimensional model that closely represents the actual location. The model consists of three different file formats:
        
        \begin{enumerate}
            \item \textbf{OBJ File (.obj)}: This file contains information about the geometry of 3D objects. Each object is defined by polygon faces, normals, curves, texture maps, and surfaces.
            
            \item \textbf{Material Library File (.mtl)}: This file defines each of the materials in the 3D model, including their color, texture, and reflection properties.
            
            \item \textbf{Portable Network Graphic Files (.png)}: Multiple PNG files are generated to store texture images for the 3D models. These files work with the MTL files to generate textures for the 3D surfaces.
        \end{enumerate}

        \begin{figure}[h]
        \centering
        \includegraphics[width=\linewidth]{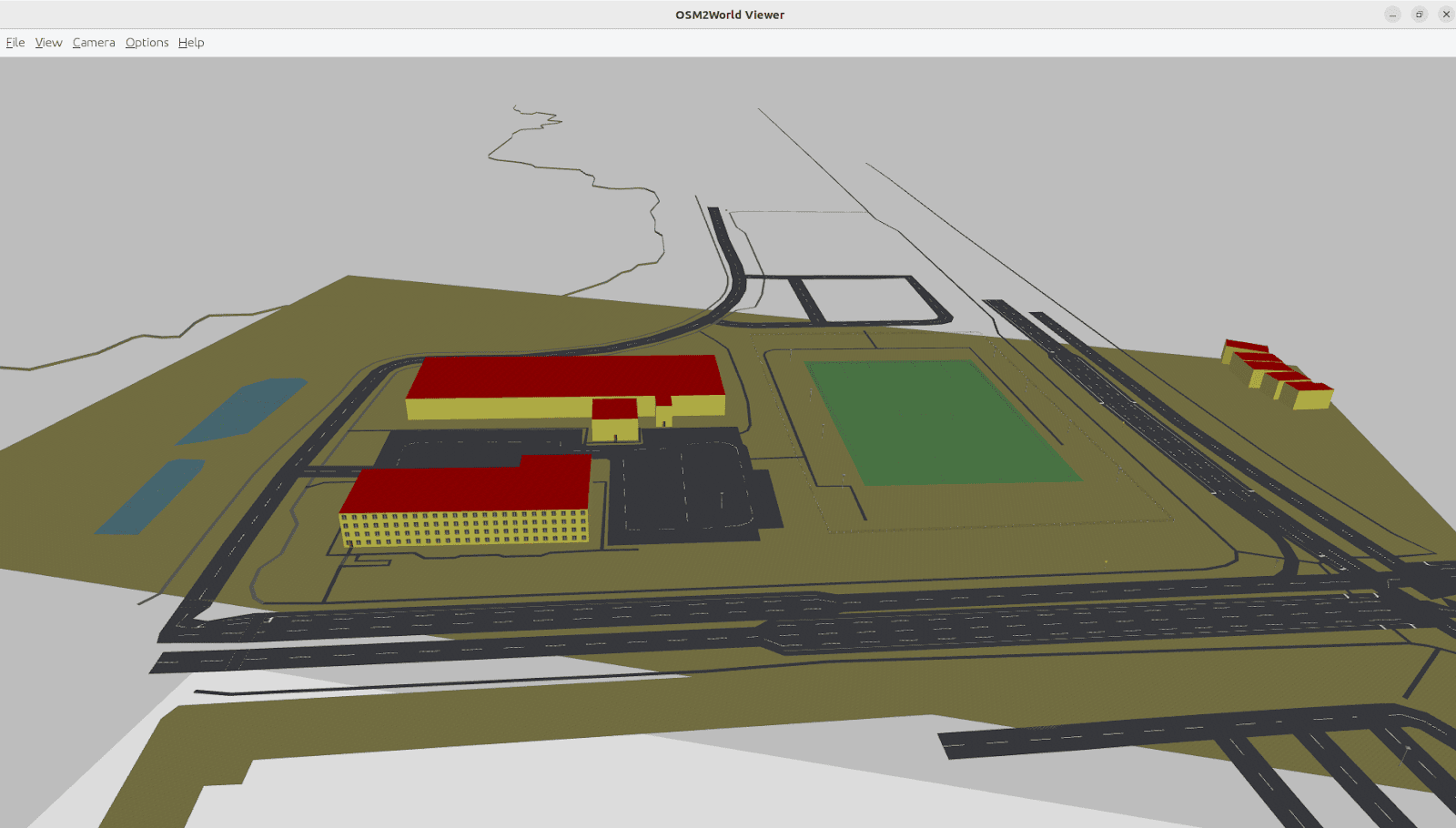}
        \caption{3D model created in the OSM2World GUI.}
        \end{figure}

        OSM2World comes preinstalled in the container and generates a 3D model shown in Figure 4, using the OSM file created earlier through its command line interface. The files are generated as one OBJ file, one MTL file, and a folder of multiple PNG files, containing pictures of the building texture, stop signs, grass, and roads.
    
        \item \textbf{CloudCompare Point Cloud Extraction}
        
        CloudCompare \cite{b12} is a 3D point cloud and triangular mesh processing software. This software is used in the container to import the 3D mesh, and export the point cloud (.pcd). This is done by first importing the 3D mesh and using the sample points feature, which calculates various dense points based on the surfaces of the mesh to generate a point cloud. This point cloud contains the set of data points in a 3D coordinate system that represent the shape of the 3D mesh.
    
        \begin{figure}[h]
        \centering
        \includegraphics[width=\linewidth]{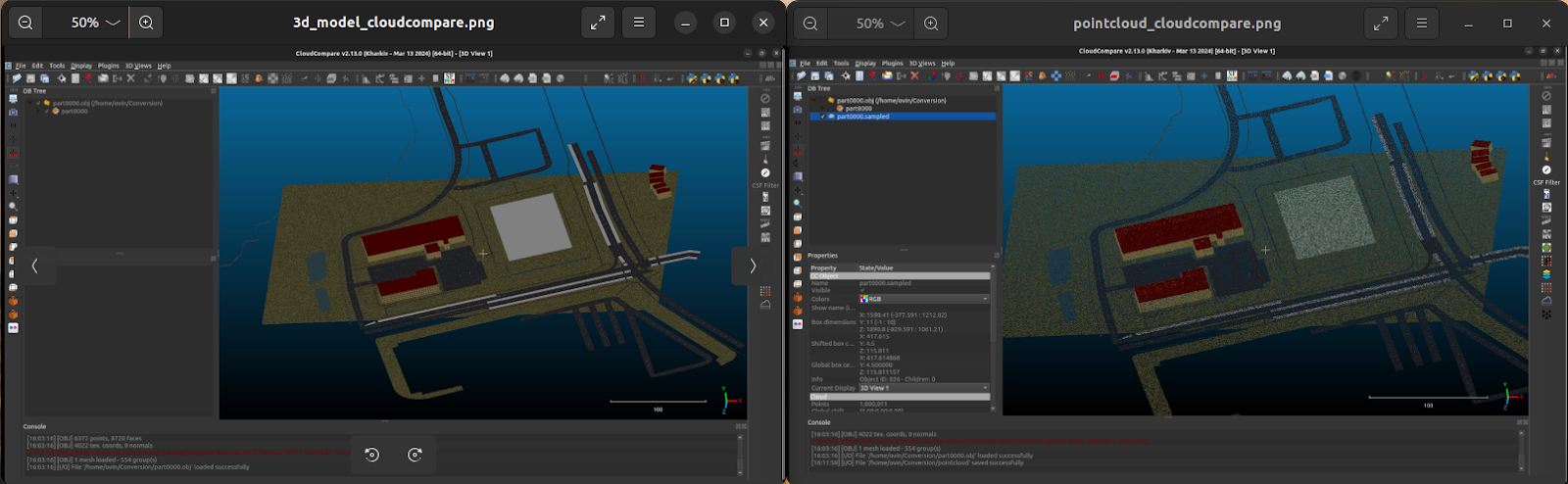}
        \caption{3D model and point cloud shown in the CloudCompare GUI.}
        \end{figure}

        This software comes preinstalled inside the Docker container, and is then used through its CLI interface to import the 3D mesh obj file generated from OSM2World and generate a point cloud of the mesh. The 3D mesh file and point cloud are shown on the left and right of Figure 5 respectively. This point cloud is a single file, represented as a PCD file.
    
        \item \textbf{Point Cloud Library (PCL) Processing}
        
        Point Cloud Library (PCL) \cite{b13} is an open source project used for point cloud processing. This library contains various features on processing point clouds, such as viewing it in a 3D space, removing outliers, connecting point clouds, creating surfaces, and many more. In this pipeline, PCL is used to fix the orientation of the point cloud from a frontal view to a top-down aerial view. It then converts the point cloud file from ASCII format to binary format. This concludes the processing of the point cloud file, making it ready for use with Autoware. This file must then be renamed to \texttt{pointcloud\_map.pcd}, due to Autoware naming conventions.
    
        \begin{figure}[h]
        \centering
        \includegraphics[width=\linewidth]{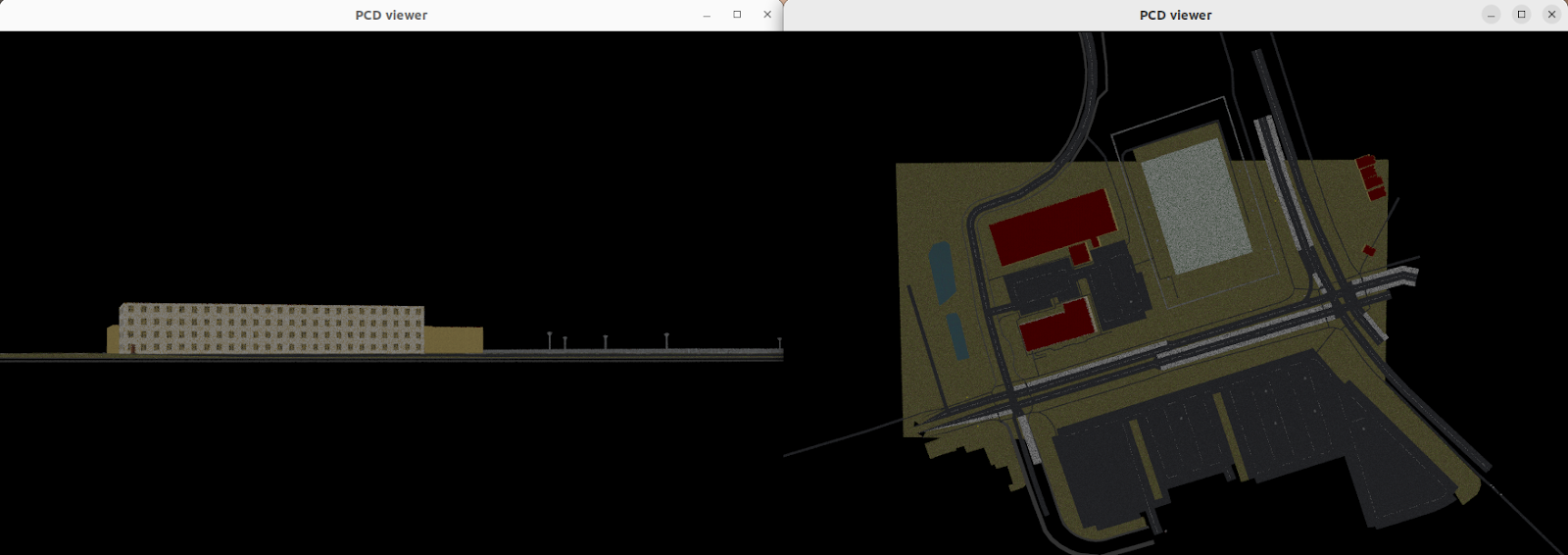}
        \caption{Orientations of point cloud before and after processing.}
        \end{figure}

        This library is used inside the Docker container, which already has the library installed. The library contains three useful functions. The first is \texttt{pcl\_viewer}, which helps to view the point cloud and its initial orientation. Upon viewing it, the orientation will be seen as an initial frontal view. This must be changed so that the initial view is a top-down view. Therefore, the next command used is \texttt{pcl\_transform\_point\_cloud} to transform the view to a top-down view. The initial view and transformed view are shown on the left and right in Figure 6 respectively. After this step, the last thing to do is to uncompress the file to binary format, using the command \texttt{pcl\_convert\_pcd\_ascii\_binary}. With these steps, the PCD file processing is completed. 
    \end{enumerate}   

    \item \textbf{Vector Map Builder}

    Vector Map Builder \cite{b14} is a tool provided by Tier IV, which is used for creating a Lanelet2 vector map, a specialized format for AV simulations. Although the resulting file uses the .osm extension, it is distinct from typical OpenStreetMap data. Lanelet2 files define road networks, lanes, and other road features essential for AV simulations. This Lanelet2 OSM file allows Autoware to run simulations on the predefined lanes. The tool has a feature to import a point cloud file, which can then be used to manually define lanes, parking lots, and parking spaces. These features can be customized as needed, but they are generally designed to conform with real-world features. After defining these features, the resulting Lanelet2 map can be exported as an OSM file.

    \begin{figure}[h]
    \centering
    \includegraphics[width=\linewidth]{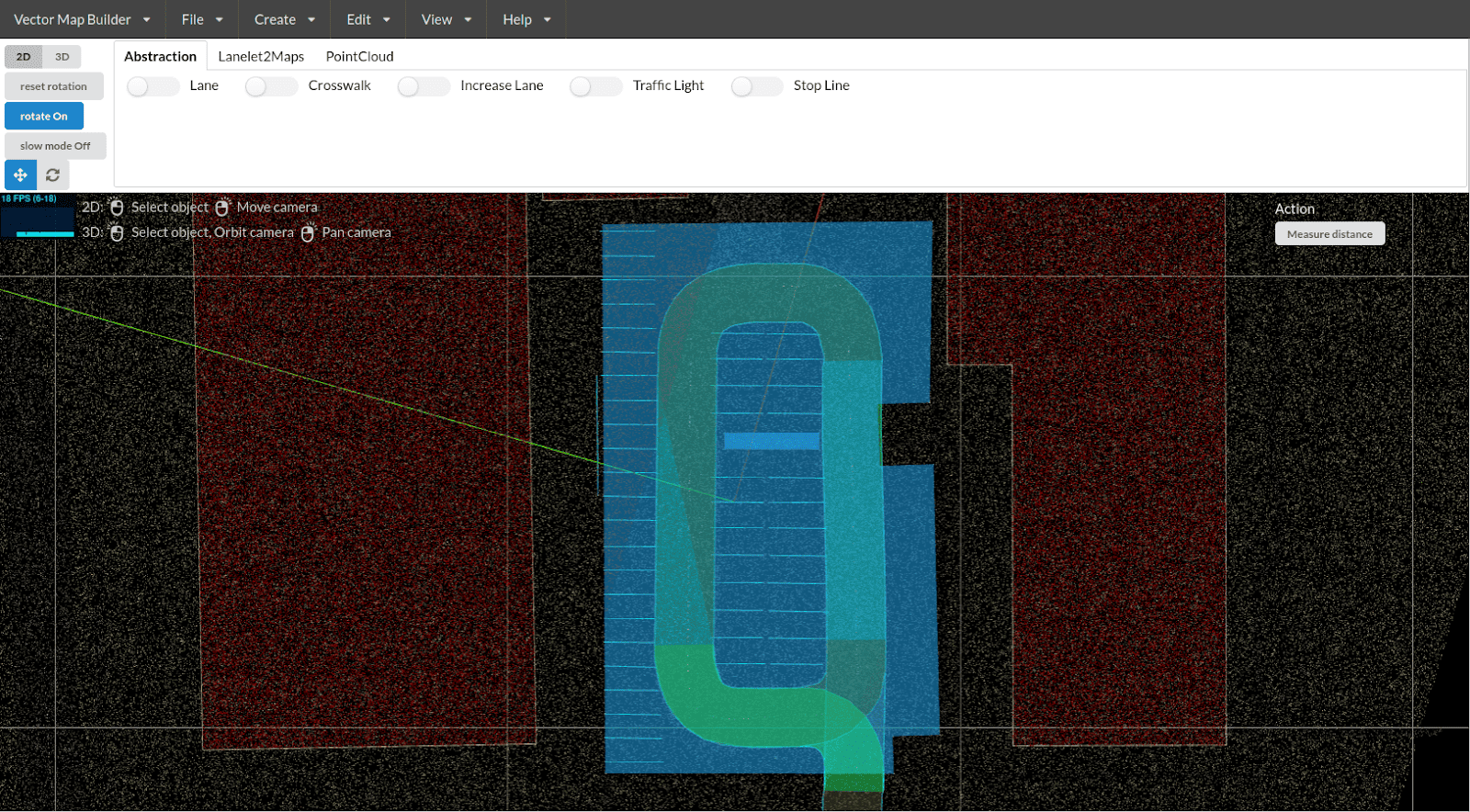}
    \caption{Lanelets and parking spaces created in VectorMapBuilder.}
    \end{figure}

    The tool can be accessed through the website, and can be used to import the point cloud file. After the import, features such as lanes and parking spots can be drawn. In the case of the SIRC parking lot, the lanes and parking spots were drawn as accurately as possible, shown in Figure 7. After completion, a \texttt{lanelet2\_map.osm} file can be exported.
    
    \item \textbf{Python Script for OSM Manipulation}
    A python script, \texttt{remove\_lat\_lon.py}, provided in \cite{b8}, was created to nullify all latitude and longitude fields from the Lanelet2 OSM file. This is necessary for functionality with the Autoware software. If the lat/long coordinates are not NULL, the lanes will malfunction and stretch into infinity in Autoware.
    
    This script is used in the Linux terminal, by invoking its command and giving it the \texttt{lanelet2\_map.osm} file as input. The script then sets all lat/long fields to NULL and outputs the updated Lanelet2 OSM file.

\end{enumerate}

\subsection{Integration}

The workflow generates three essential files: a Lanelet2 OSM file, a PCD file, and 3D mesh files (OBJ, MTL, and PNG). To ensure compatibility with Autoware, the Lanelet2 OSM file must be named \texttt{lanelet2\_map.osm}, and the PCD file should be named \texttt{pointcloud\_map.pcd}. These files can then be imported into Autoware and AWSIM, as detailed below:

\begin{enumerate}
    \item \textbf{Autoware}
    
    Autoware requires the Lanelet2 OSM file, and the PCD file. It is then launched with a specific ROS2 launch command with the map path argument pointing to the two files. 

    \begin{figure}[h]
    \centering
    \includegraphics[width=\linewidth]{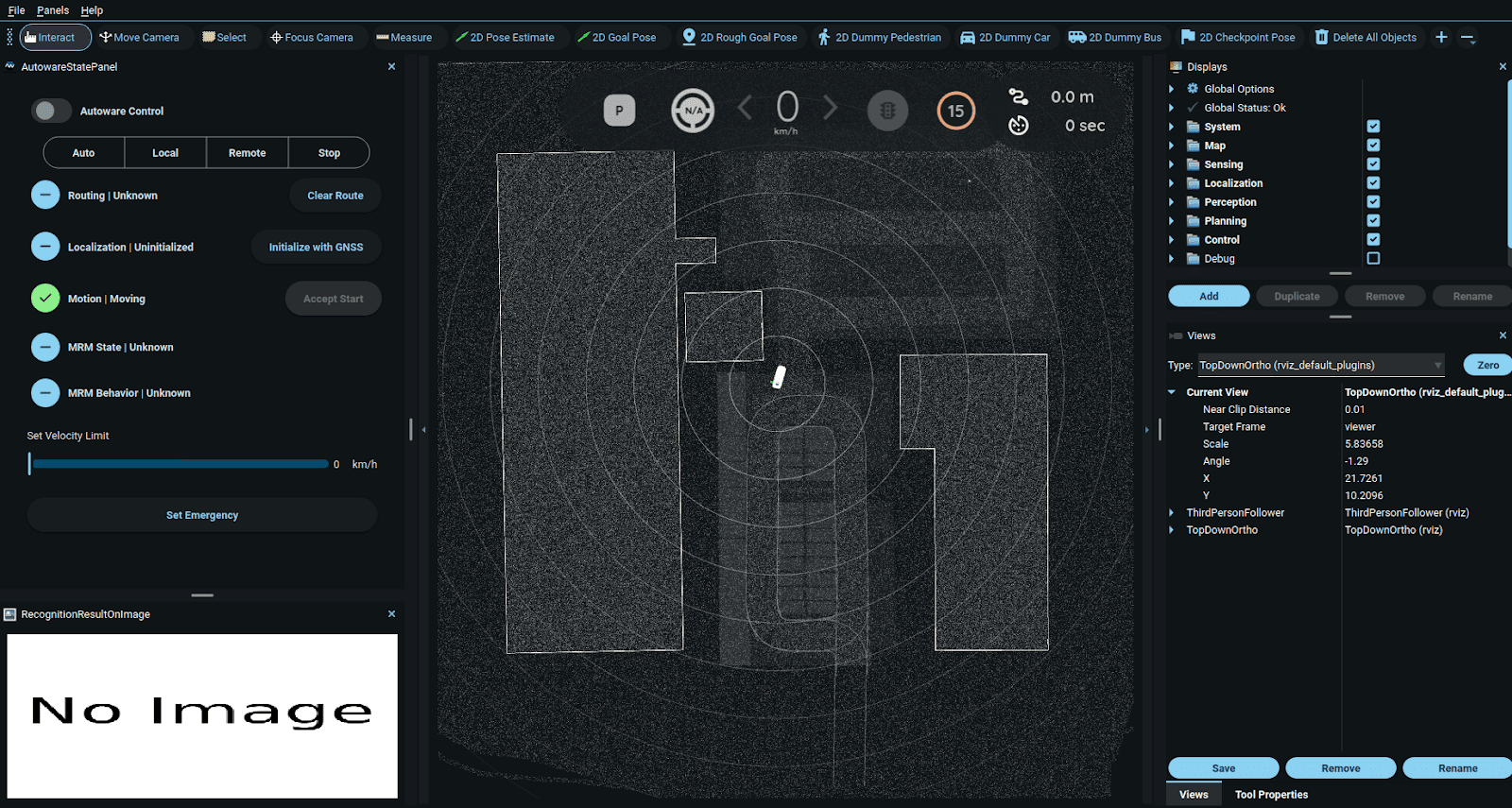}
    \caption{Lanelet2 map and point cloud imported into Autoware.}
    \end{figure}
    
    Figure 8 shows the correct loading of the two files. 
    
    \item \textbf{AWSIM}
    
    AWSIM requires the 3D mesh and Lanelet2 OSM file to be imported in. The 3D mesh then needs some additional steps done to start working. Some scripts have to be added which define the 3D mesh as Mesh Colliders, so that they can be interacted with in the simulation. The 3D mesh file also has to have the read/write field enabled. Lastly, the Lanelet2 OSM file is loaded and aligned with the simulation environment to synchronize with Autoware. Figure 9 shows the correct loading of the 3D files and the Lanelet2 map.

    \begin{figure}[h]
    \centering
    \includegraphics[width=\linewidth]{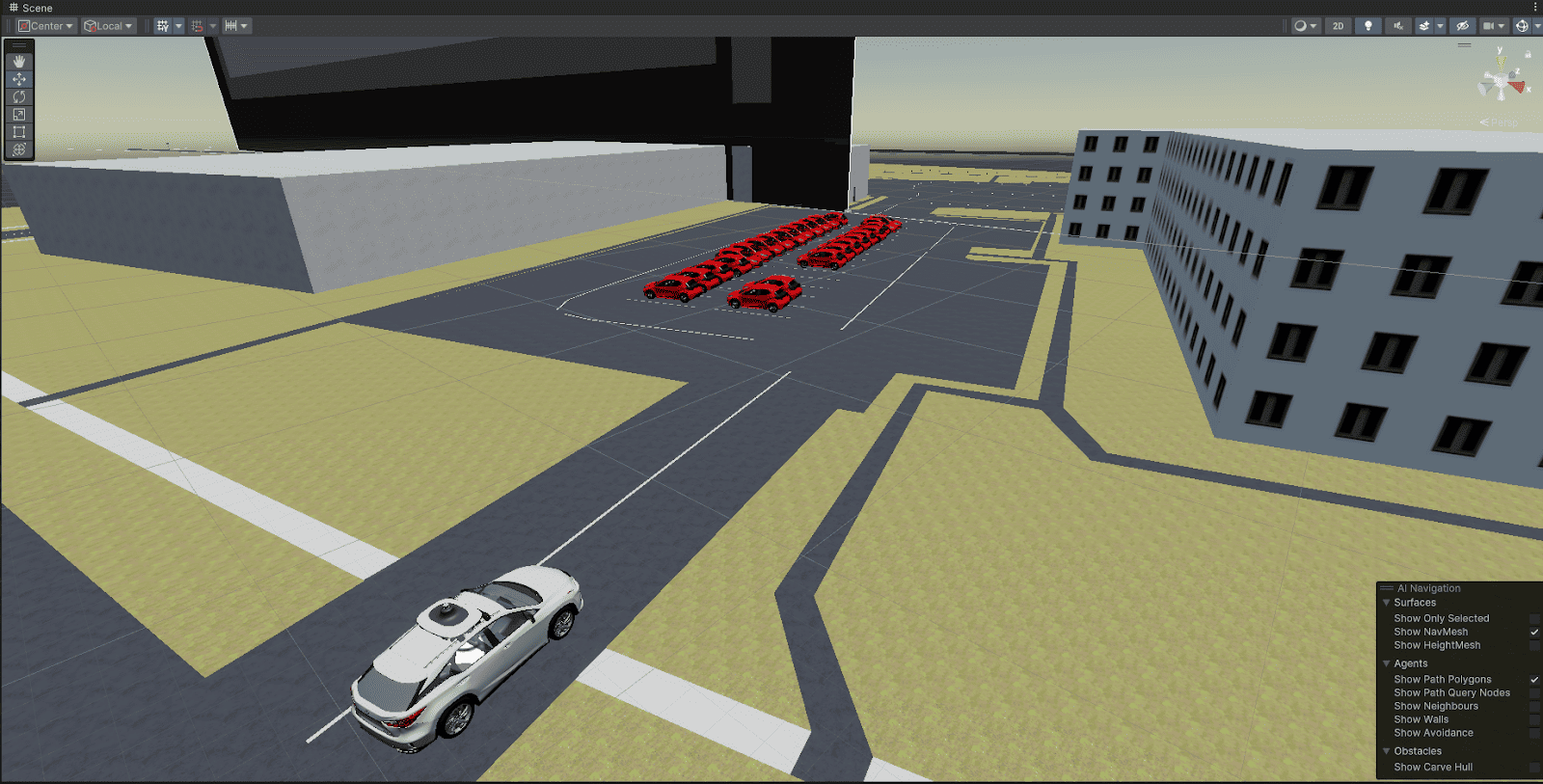}
    \caption{Lanelet2 map and 3D model imported into AWSIM.}
    \end{figure}

\end{enumerate}

\section{Results}

After importing all the files of the newly created map, both Autoware and AWSIM are able to load the SIRC parking lot environment. To get both synced, AWSIM is run first, and then Autoware afterwards. In AWSIM, the ego vehicle is correctly spawned and activated inside the environment, with all its sensors functioning correctly. In Autoware, the ego vehicle is correctly localized to the position of the ego vehicle by receiving the location from AWSIM. The initialization of AWSIM and Autoware are shown in Figure 10. 

\begin{figure}[h]
\centering
\includegraphics[width=\linewidth]{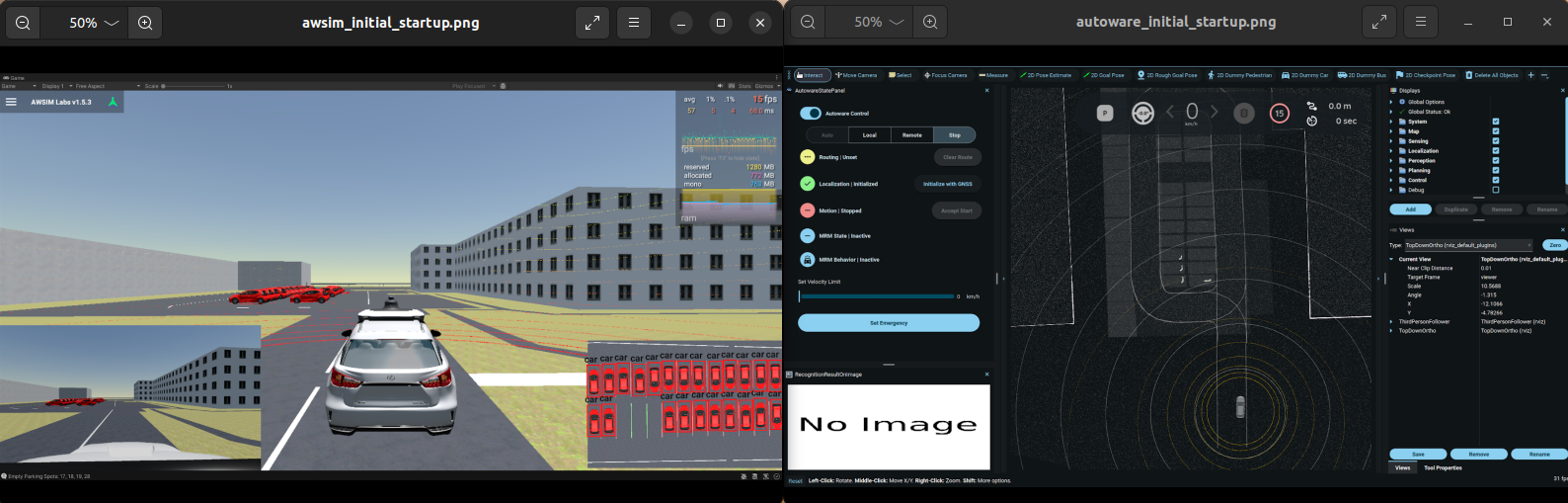}
\caption{Initial startup of AWSIM and Autoware.}
\end{figure}

\begin{figure}[h]
\centering
\includegraphics[width=\linewidth]{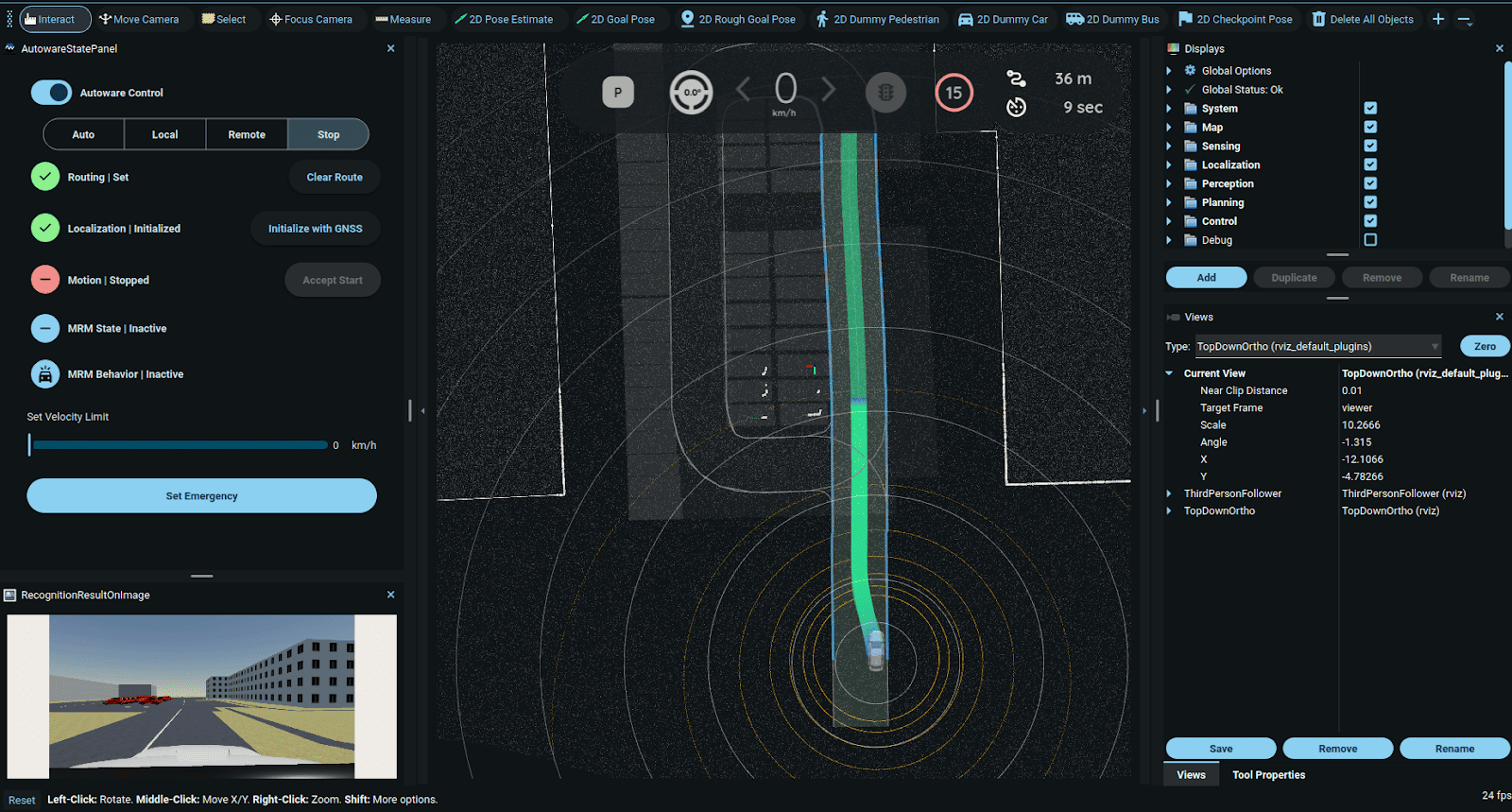}
\caption{Goal pose selected and route calculated.}
\end{figure}

\begin{figure}[h]
\centering
\includegraphics[width=\linewidth]{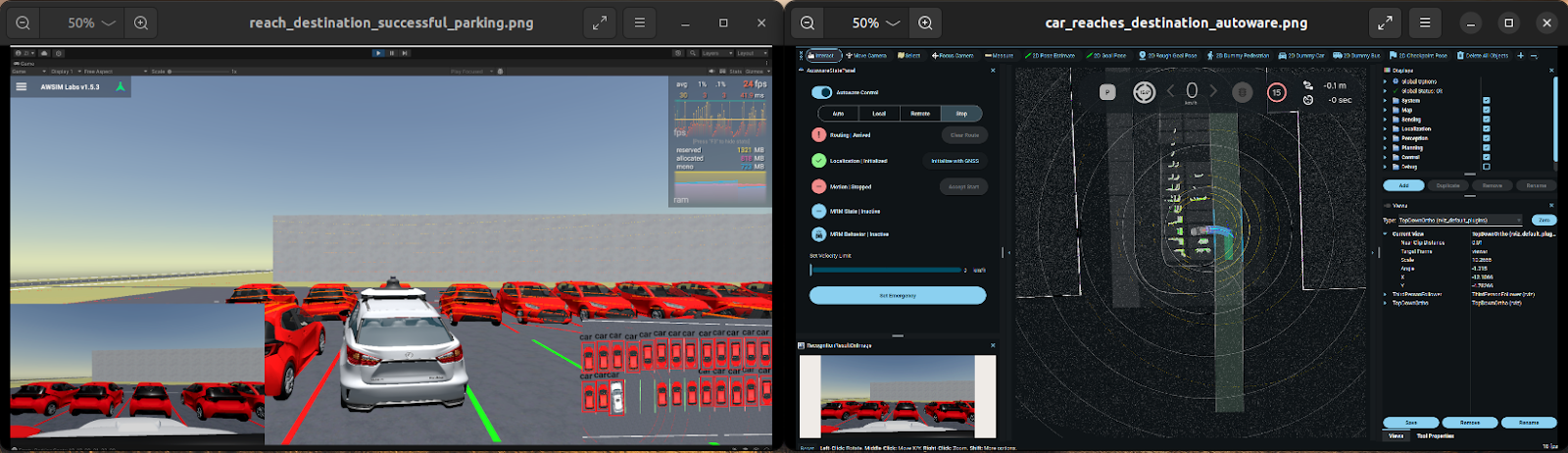}
\caption{Car reaches the destination in AWSIM and Autoware.}
\end{figure}

After setting a goal pose inside a parking spot for the vehicle, which is shown in Figure 11, and activating the autonomous mode, both vehicles in AWSIM and Autoware accurately mimic each other and reach the destination correctly, which are shown in Figure 12.

The successful completion of route planning and parking demonstrates the map’s effective integration into both simulation platforms, highlighting its accuracy and the ego vehicle's proper localization and navigation in AWSIM and Autoware. These results show the workflow’s capability to support real-world applications and test AVP systems in simulation environments.

\section{Discussion}

With this workflow, testing can be done in any outdoor environment that is available on OpenStreetMap. AWSIM has the ability to generate pedestrians and traffic, and also relays all this information back to Autoware. In any scenario, whether it is simple driving, or parking, Autoware can be used to test them. In order to deploy Autoware in real life, an alternative must be used for generating the PCD and Lanelet2 maps. This is because the 3D model generated by OSM2World is not perfect. The buildings are not true to reality. For example, the SIRC parking lot contains a soccer dome, and in OSM2World, this dome is represented as a simple rectangular building. However, in the case of real-life deployment, SLAM technologies can be used to generate the perfect and accurate point cloud.

Although the 3D models generated by OSM2World have limitations in geometric accuracy, the workflow remains highly practical compared to other HD map generation methods such as HDMapNet or CARLA-based pipelines. Unlike those approaches, which rely on real-world sensor data and manual annotation, this workflow generates all required map components offline using only open-source tools and OSM data. This makes it especially suitable for simulation use, offering accessibility and ease of deployment in low-resource environments. Although formal evaluation metrics such as runtime or accuracy comparisons are not presented, the workflow's successful integration with AWSIM and Autoware demonstrates its effectiveness for early-stage research and prototyping.

\section{Conclusion}

In this paper, a custom workflow was presented, which was designed to simplify the creation of maps for use with AWSIM. While primarily developed by AWSIM, this workflow can potentially be adapted for use with other simulators, though this aspect was not explored in detail. The workflow was developed within the context of an AVP project using Autoware, addressing a critical gap in the availability of simulation-ready environments for testing AV technologies.

Accurate and adaptable maps are essential for AV development. However, creating such maps can often be difficult and resource-intensive. Many existing workflows rely on significant computational resources or are tied to specific simulators, limiting their flexibility for developers. Moreover, documentation for creating custom maps can often be difficult to follow, complicating the process of integrating real-world locations into simulations. This workflow addresses these challenges by using lightweight tools to generate 3D mesh files, point cloud data files, and Lanelet2 files from OSM data, making it applicable to any location available on OSM.

HD maps are vital for testing AVs in simulated environments before real world deployment. These maps offer centimeter-level accuracy for road layouts, lane markings, and traffic infrastructure, supporting localization, perception, and planning tasks. The workflow demonstrated in this paper enables the creation of HD maps for use in 3D simulation engines, including AWSIM, which is compatible with Autoware. The simulated maps are not only valuable for virtual testing but also serve as the foundation for real-world deployment, ensuring a seamless transition from simulation to real-world navigation.

The methodology outlined in this paper, ranging from extracting data from OpenStreetMap to processing it through Docker containers for map generation, was used to create a functional 3D map of Ontario Tech University’s SIRC parking lot. This map was successfully tested in both Autoware and AWSIM simulations, demonstrating the workflow’s potential for use in a variety of real-world environments available on OpenStreetMap. Future work will focus on improving model accuracy, incorporating SLAM technologies, and optimizing the workflow for broader simulator compatibility. Additionally, exploring more flexible handling of latitude and longitude values could allow for better control over the nullification process and further enhance map generation accuracy.

\section*{Acknowledgment}
We would like to extend our gratitude to our friends and team members, Waddah Saleh and Abdullah Waseem, for their encouragement and support throughout the project. Additionally, we would like to thank our research group for their continuous guidance and valuable insights during the course of this work.

\vspace{12pt}

\end{document}